\documentclass[a4paper,twoside]{article}

\usepackage{epsfig}
\usepackage{calc}
\usepackage{amssymb}
\usepackage{amstext}
\usepackage{amsmath}
\usepackage{amsthm}
\usepackage{multicol}
\usepackage{pslatex}

\usepackage{natbib}
\let\bibhang\relax

\usepackage{apalike}
\usepackage{algorithm}
\usepackage{algpseudocode}
\usepackage{subfig}
\usepackage[euler]{textgreek}
\usepackage{amssymb}
\usepackage{graphicx}
\usepackage{csquotes}
\usepackage{enumerate}
\usepackage{latexsym}
\usepackage{url}
\usepackage{subfig}
\usepackage{array}
\usepackage{spverbatim}
\usepackage{comment}
\usepackage{eurosym}
\usepackage{booktabs}
\usepackage{adjustbox}
\usepackage{subfig}
\usepackage{svg}
\usepackage{multirow}
\usepackage{adjustbox}
\usepackage{xcolor}

\usepackage{float}

\usepackage{SCITEPRESS}     

\begin{document}

\title{PronounFlow: A Hybrid Approach for Calibrating Pronouns in Sentences}

\author{\authorname{Nicos Isaak\orcidAuthor{0000-0003-2353-2192}}
	\email{nicosi@acm.org}
	\affiliation{Computational Cognition Lab, Cyprus}}


\abstract{Flip through any book or listen to any song lyrics, and you will come across pronouns that, in certain cases, can hinder meaning comprehension, especially for machines. As the role of having cognitive machines becomes pervasive in our lives, numerous systems have been developed to resolve pronouns under various challenges. Commensurate with this, it is believed that having systems able to disambiguate pronouns in sentences will help towards the endowment of machines with commonsense and reasoning abilities like those found in humans. However, one problem these systems face with modern English is the lack of gender pronouns, where people try to alternate by using masculine, feminine, or plural to avoid the whole issue. Since humanity aims to the building of systems in the full-bodied sense we usually reserve for people, what happens when pronouns in written text, like plural or epicene ones, refer to unspecified entities whose gender is not necessarily known? Wouldn't that put extra barriers to existing coreference resolution systems? Towards answering those questions, through the implementation of a neural-symbolic system that utilizes the best of both worlds, we are employing PronounFlow, a system that reads any English sentence with pronouns and entities, identifies which of them are not tied to each other, and makes suggestions on which to use to avoid biases. Undertaken experiments show that PronounFlow not only alternates pronouns in sentences based on the collective human knowledge around us but also considerably helps coreference resolution systems with the pronoun disambiguation process.}

\onecolumn \maketitle \normalsize \setcounter{footnote}{0} \vfill

\section{Introduction}
``Midnight, not a sound from the pavement... Has the moon lost her memory..." We start this section with the initial lyrics of \textit{Memory}\footnote{\url{https://en.wikipedia.org/wiki/Memories_(Barbra_Streisand_album)}}, a famous 80s song performed by Barbra Streisand and written by Andrew Lloyd Webber, T. S. Eliot, and Trevor Nunn. For anyone listening or reading for the first time, questions pop up in their mind about whom or what the pronoun \textit{her} refers to ---in the snippet of context \textit{her Memory}. Some might argue that the definite pronoun \textit{her} refers to the moon, as it is feminine according to their collective knowledge, while others to a woman not yet introduced. Other cultures that view the moon as a more neutral entity might argue in favor of the pronoun \textit{it}, while others in favor of a more epicene one like \textit{xe, ey, or ze} \citep{lauscher2022welcome}.

Nevertheless, the issue that our statements raise is that pronoun resolution is sometimes challenging even for humans, but especially hard for machines, as it relates to reading comprehension, which depends on explicit and implicit statements the writer uses to transfer their message along with the commonsense knowledge already known by the reader \citep{mckoon1992inference}. Disambiguating pronouns signify the use of semantic constraints to retrieve the actual referents or antecedents \citep{corbett1983pronoun}. It is well-studied that humans, when tackling pronouns in sentences, see invisible structure, not linear order, which is necessary to understand how pronouns can refer to something \citep{adger2019language}. 

In this work, we argue that having tools able to tie pronouns and entities would not only considerably help existing co-reference resolution systems perform a better assignment of pronouns to appropriate antecedents, but also help towards the development of novel systems able to parse sentences with pronouns and make suggestions for alternate ones, according to the existing collective human knowledge found on the WWW, where a version of it exists in already developed language models \citep{devlin2018bert,liu2019roberta}.

If you want to know more about our approach, stay on this journey, which is untangled in the following sections. We start with our Motivation section by shedding light on how various groups of people use pronouns according to their beliefs and biases. Following up, we explain what Pronoun Disambiguation is all about and how existing systems handle it. Next, we present our Methodology section, showing how our system, PronounFlow, utilizes existing tools to calibrate pronoun usage in sentences. In our Experimental section, we show the feasibility of our approach with two undertaken experiments, and finally, in the last section, we conclude with our findings and discuss the implications of our research along with our thoughts on future work.

\section{Motivation}
Since the late 50s, a handful of AI pioneers set the pace for developing congenial machines that one day could act, behave, or even think like humans \citep{mccarthy2006proposal}. Following this pace, especially with the advent of neural networks, we are on the cusp of a new revolution where everybody uses AI to tackle various challenges, like answering questions, summarizing, or translating text \citep{lewis2019bart}. 

One challenge relating to all the above is the disambiguation of pronouns, that is, the pronoun assignment to proper antecedents such as nouns or complete noun phrases, a crucial action for figuring out the meaning in written or verbal conversation \citep{caramazza1977comprehension,corbett1983pronoun}. Flip through any book, or listen to any conversation, and you will encounter various pronouns. In this sense, while reading something, figuring out the actions or, more simply, who did what to whom is essential to interpret its meaning without any misunderstandings \citep{michael2009reading,marcus2019rebooting}. Simply put, for pronoun resolution, one needs to encode its integrated clause and find other related clauses to find the actual referent.

Consequently, developing systems to disambiguate pronouns, especially in well-structured sentences, is challenging and troublesome \citep{bender2015establishing}. This process is even more complicated when we know that pronouns or even word adaptations and abbreviations are in everyday usage as they help people express themselves everywhere, from their daily social events up to social media where text limitations apply \citep{al2020influence,strain2016perceptions}. This is more prominent with technology shaping our future, as it influences our speaking and language abilities \citep{al2020influence}. Commensurate with this, younger people adapt their behavior based on their daily interactions on social networks or even change their writing skills based on imposed Twitter-text limitations. 

It seems that the world is changing at a fast pace along the pronoun usage, too, based on which a more gender-fair language is used \citep{lindqvist2019reducing}. In this sense, different strategies have been proposed to impose a new inclusive era of gender-neutral pronoun usage rather than traditionally used ones \citep{lindqvist2019reducing}. Many studies are challenging the traditionally binary notion of gender in order to reduce male bias, according to which neutral words are mostly associated with male entities \citep{lindqvist2019reducing,merritt2006gender}. Subsequently, various academic and professional organizations instruct their members to avoid using masculine or even feminine pronouns in sentences and instead start using neutral gender ones, such as they and theirs. Similarly, neo or epicene pronouns emerged as a neologism for pronouns used by non-binary, transgender, or gender-nonconforming people. For example, epicene pronouns like xe, ey, or ze can be used instead of gendered ones like he, she, or even they-them-their, which do not seem to gain universal acceptance as non-binary pronouns.

However, these adaptations impose a new era of problems for academicians or educators, making it difficult for them to put barriers between formal and informal English to where the threshold begins \citep{al2020influence,isaak2023neuralsymbolic}. On another, by alternating pronouns or pluralizing, people might introduce ambiguity in cases with sentences with more than one plural referent \citep{madson2006alternating,madson2001readers}. From this point of view, pronoun resolution might be difficult in non-gendered languages like modern English, especially when people try to alternate by using masculine, feminine, plural, or neopronouns to avoid the whole issue, which makes the building of intelligent machines even harder.

The issue that our work raises is the following. With new pronouns in the scene, further problems arise for developing or utilizing existing coreference resolution systems in the pronoun disambiguation process, making matching pronouns with entities even harder. In this line of work, scant attention has been paid to building systems that would considerably help or aid towards a better coreference resolution analysis. According to current collective human knowledge found in commonly used training data across the WWW, these systems could act as parsers of sentences to reduce the usage of uncommonly used pronouns and reduce barriers to comprehending meaning. Consequently, these parsers could parse sentences prior to any coreference resolution task. 

According to \citet{chang2023speak}, there are systematic biases in what kind of material a language model is trained on, similarly reflected from search engines around us. Additionally, being able to alternate pronouns based on a pronoun-gender agreement would suffice to help other kinds of systems, like Neural Machine Translation services \citep{saunders2020neural,icaart20}. Given that it is widely common for the NLP community to use out-of-the-box models or fine-tune already trained ones to tackle almost everything \citep{howard2018universal}, here we employ a goal-oriented strategy that helps systems achieve a better pronoun resolution based on readily available information \citep{corbett1983pronoun}. To that end, we blended a simplified tool developed for pronoun resolution in well-structured sentences with a trained language model that natively predicts missing masked tokens from a hidden text \citep{isaak2021blending,icaart20}. 

Given that, at the time of writing, the majority of language models are trained on vast amounts of \textit{pre-neopronoun} data, this might lead to an added layer for better co-reference resolution, where language models are used to replace pronouns in sentences with the ones selected themselves based on the collective human knowledge found in their training data. Additionally, knowing that pronoun resolution is easier with gendered pronouns \citep{brandl2022conservative}, PronounFlow was pivoted to make decisions based on a gendered entity-pronoun matching approach. To date, this is the first published work to report results on the feasibility of this approach.

\section{Pronoun Disambiguation}
Pronoun disambiguation \citep{corbett1983pronoun} or similar tasks like the well-known Winograd Schema Challenge (WSC) \citep{levesque2012winograd,levesque2014our} refer to resolving pronouns in sentences or phrases. The basic idea behind these challenges is the building and probing of systems' capabilities in resolving definite pronouns, especially in cases where basic shallow techniques are not proper (see a PDP in Table \ref{Comparing}). Since the late seventies, \citet{hobbs1978resolving} stated that pronoun resolution would come without effort after we do everything else. Subsequently, it is believed that having systems able to tackle pronouns in sentences will advance and push the field of AI toward building machines with human-like reasoning abilities.

Especially the WSC, a more advanced test referring to resolving pronouns in well-structured twin sentences (schemas) where two referents are of the same number and gender, could be seen as a novel litmus test to push the AI research community even further without using statistical techniques. It is worth noticing that well-constructed schemas differ only in a \textit{special word}, which, when replaced, the referent of the pronoun also changes. Although this is hard to measure, systems able to solve Winograd schemas should be able to demonstrate their human-like reasoning abilities. However, in retrospect, with the development of opaque language models that can memorize almost everything \citep{chang2023speak}, we have developed systems that tackle schemas, sometimes even better than humans, though this is still debatable \citep{elazar2021back,kocijan2023defeat}.

Overall, various systems disambiguate pronouns based on neural networks, feature-based, or knowledge-based approaches.

To illustrate, in the category of knowledge-based approaches, \citet{kn:sharma2015towards} utilized Answer Set Programming (ASP) to build a parser that learns to do pronoun resolution via the learning of semantically and structurally similar examples from English sentences. Similarly, \citet{schuller2014tackling} disambiguates pronouns via encompassed dependency parsers that analyze the structure of sentences with manually created background knowledge graphs, while \citet{hong2021tackling} do the same through an ensemble of knowledge-based reasoning and machine learning techniques. \citet{Isaak2016}, built Wikisense, a commonsense conclusion engine that solves cases of pronoun problems based on a transparent policy, in the sense that, we can easily understand why and how it takes certain decisions by examining its beliefs and reasoning.

Regarding feature-based approaches, \citet{fahndrich2018marker} resolved definite pronouns in sentences using graphs from sources like WordNet and Wikipedia, all merged with syntactical information from dependency parsers. \citet{emami2018knowledge}, via queries that capture the predicates in examined sentences, search the WWW to retrieve relevant snippets, which are then parsed and filtered to find the correct pronoun antecedents.

In the most prominent area of neural networks, researchers maximize the efficiency of their approaches through language models, albeit without transparency, as these models are often characterized as opaque \citep{marcus2019rebooting}. In this line of work, researchers probe language models such as unified ensembles of LSTMs or transformers such as BERT and GPT \citep{radford2018improving}, which indirectly encode human knowledge found in various corpora, to solve various versions of pronoun resolution problems.  \citet{trinh2018simple} did that by replacing definite pronouns with their respective possible referents and searching the models. \citet{klein2019attention} tackled pronouns based on context-aware word embeddings learned from BERT's attention maps \citep{devlin2018bert}. Similarly, \citet{kocijan2019surprisingly} utilized the BERT language model on its innate masked token prediction ability by fine-tuning it further in well-structured sentences to learn which pronoun target was the best replacement for each masked entity. Additionally, many other researchers approach the problem through other models like RoBerta, Text-to-Text Transfer Transformer, or hybrid neural network approaches \citep{raffel2020exploring,https://doi.org/10.48550/arxiv.1907.10641,he2019hybrid}.

\begin{table*}[h!]
	{\caption{Pronoun Disambiguation Problems: An example of a simple pronoun disambiguation problem on the first three rows, while on the last three, a schema from the Winograd Schema Challenge. Examples were taken from the Commonsense and Reasoning website\footnote{\url{http://commonsensereasoning.org/}}.}\label{Comparing}}
	\begin{adjustbox}{}

		
		\begin{tabular}{|c|c|c|}
			\hline 
			\multirow{5}{*}{PDP} & \multirow{3}{*}{Phrase} & While Nancy and Ellen counted the silverware, Mrs. Smith \tabularnewline
			&  & hastened upstairs. In a few minutes she returned and one look \tabularnewline
			&  & at her stricken face told the girls that the precious map was gone.\tabularnewline
			\cline{2-3} \cline{3-3} 
			& Pronoun to disambiguate & \textit{her} stricken face\tabularnewline
			\cline{2-3} \cline{3-3} 
			& Possible referents & a) Nancy (b) Ellen (c) Mrs. Smith \tabularnewline
			\hline 
			\multirow{7}{*}{WSC} & \multirow{2}{*}{1st Sentence} & The town councilors refused to give the demonstrators \tabularnewline
			&  & a permit because they feared violence. \tabularnewline
			\cline{2-3} \cline{3-3} 
			& Question & Who feared violence?\tabularnewline
			\cline{2-3} \cline{3-3} 
			& \multirow{2}{*}{2nd Sentence} & The town councilors refused to give the demonstrators\tabularnewline
			&  & a permit because they advocated violence.\tabularnewline
			\cline{2-3} \cline{3-3} 
			& Question & Who advocated violence?\tabularnewline
			\cline{2-3} \cline{3-3} 
			& Answers & the town councilors, the demonstrators \tabularnewline
			\hline 
		\end{tabular}
\end{adjustbox} 
\end{table*}

\section{Methodology}
In this section, we present PronounFlow's architecture with the modules it consists of. When language models are trained on vast swaths of unsupervised text, they utilize a conceptually collective dense network of rules to tackle various challenges. Most commonly, through transfer learning, a flurry of language models, like BERT, can be fine-tuned on downstream tasks to solve challenges they were never trained for. From this standpoint, the new orthodoxy in AI needs only two things, access to small amounts of training data for fine-tuning and a few dedicated GPUs. Here, like in many other works, we utilize BERT, one of the most well-known foundational models ever developed and trained on two known tasks: a mask-token and a sentence prediction task \citep{devlin2018bert}. In the former task, the model is trained on predicting missing tokens in English sentences, while in the second on predicting if a sentence can be considered the follow-up of a previous one. 

Especially mask-token prediction, usually referred to as masked auto-encoding, is based on the simplified task of predicting a hidden portion of text based on its surroundings. The idea of masked auto-encoding is so well-defined in the area of NLP that it is now applied to other tasks like vision to predict missing parts of images \citep{he2022masked}.

By focusing on the mask-token prediction task, we built PronounFlow, a robust system that calibrates pronoun usage in sentences. Our approach is based on the idea of replacing pronouns in sentences based on partial observation of their surroundings. However, knowing that large language models often hallucinate facts, leading the research community to use other mostly external tools to keep them in balance \citep{chang2023speak,icaart23}, to build our system, we employed a neural-symbolic approach that utilizes components from both Classic and Modern AI. Put simply, given a sentence that contains pronouns, we filter its words based on their part of speech, match their pronouns with entities based on their gender to find any inconsistencies, and finally, replace any inconsistent pronouns with consistent ones returned by a language model. In short, below, we apply a simple but effective plug-and-play strategy, based on which suggested pronouns replace every other one for which no gender agreement exists. As an innate transparent, non-opaque hybrid system, it always tries to explain why it took certain decisions. Conversely, if it cannot assign pronouns to entities using its symbolic part, it alternates pronouns solely through its neural part —more on its innate mechanisms in the following paragraphs.

\subsection{Parser}
\begin{figure*}[h!]
	\centerline{\includegraphics[width=\textwidth]{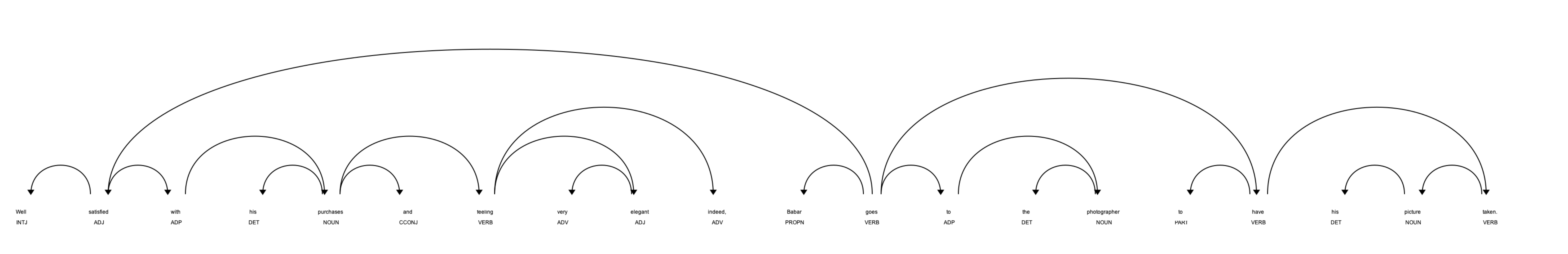}}
	\caption{A Parsed Sentence: SpaCy displays each word's part-of-speech along their relations to other words.}
	\label{spacyParsed}
\end{figure*}

\begin{figure*}[h]
	\centerline{\includegraphics[width=\textwidth]{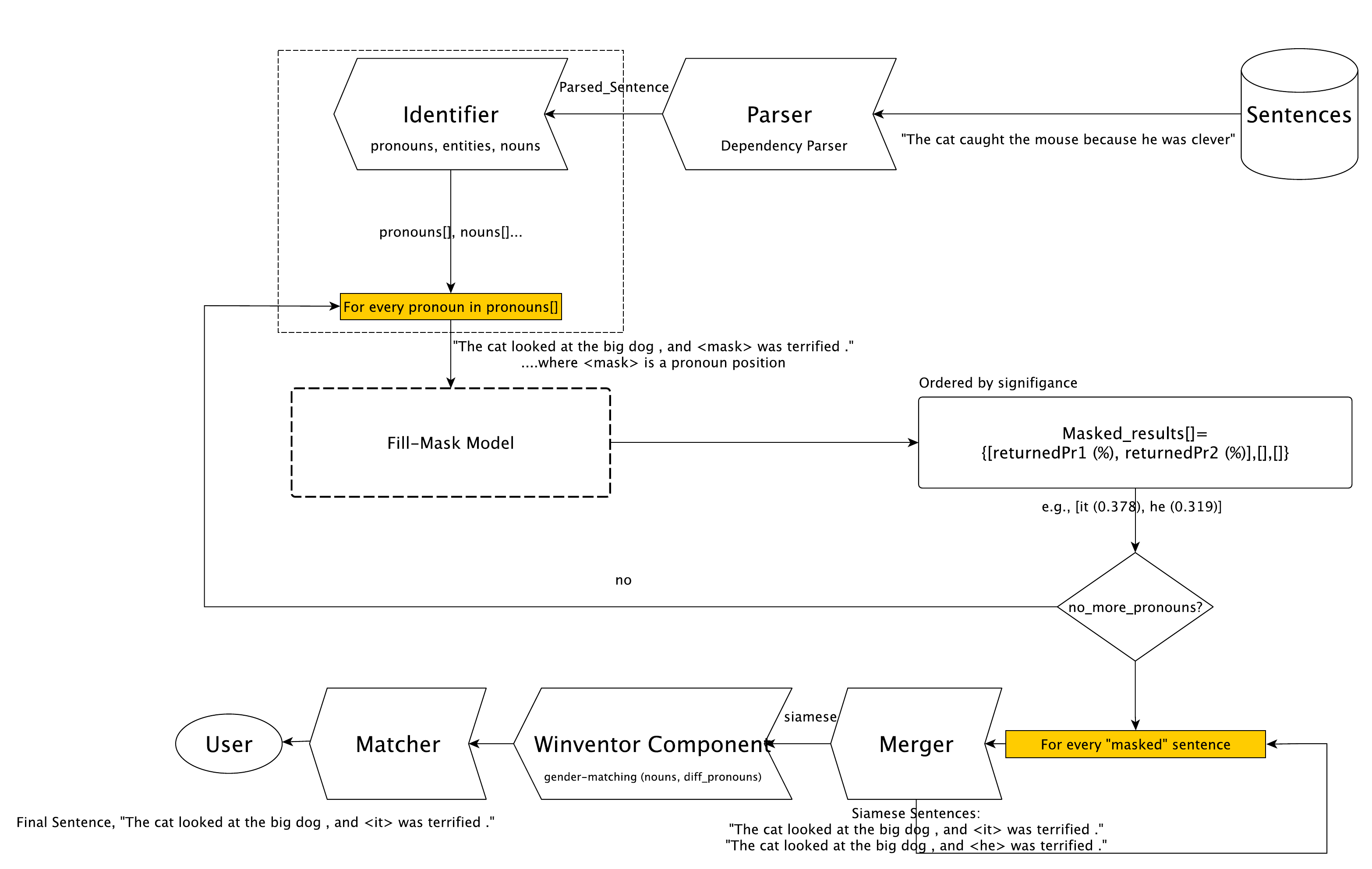}}
	\caption{PronounFlow: A Hybrid Approach for Calibrating Pronouns in Sentences.}
	\label{methodologyfig}
\end{figure*}

Dependency parsers are everywhere, as they are essential in identifying the structure and meaning of sentences \citep{de2008stanford,nivre2005dependency,choi2015depends,Isaak2016,icaart23}. In accordance with the above, below, we use spaCy, one of the fastest out-of-the-box dependency parsers, part-of-speech taggers, and named entity recognizers around us (\url{https://spacy.io}). Through spaCy, PronounFlow analyses the structure of sentences to find entities such as capitals, countries, cities, institutions, and agencies along with every word part of speech, such as pronouns, verbs, nouns, and adjectives (see a parsed example in Figure \ref{spacyParsed}, and \textit{Parser} in Figure \ref{methodologyfig}). On a second front, spaCy is utilized by the Winventor Component \citep{icaart20,isaak2021blending} to find relations between pronouns, nouns, and noun phrases (see \textit{Winventor} in Figure \ref{methodologyfig}).

\subsection{Identifier}
This component takes place after each sentence's parsing (see \textit{Upper-left corner} in Figure \ref{methodologyfig}). The Identifier's purpose is three-fold. First, for each examined sentence it filters out sentences that do not contain pronouns. Second, it categorizes each word based on their part-of-speech and entity type. Third, for every given pronoun position, it masks its place to prepare the sentence for our mask-fill language model. For instance, in Figure \ref{spacyParsed}, for the sentence, ``Well satisfied with his purchases and feeling very elegant indeed, Babar goes to the photographer to have his picture taken.", the result would be two \textit{masked} sentences: i) ``Well satisfied with $<MASK>$ purchases and feeling very elegant indeed, Babar goes to the photographer to have his picture taken." ii) ``Well satisfied with his purchases and feeling very elegant indeed, Babar goes to the photographer to have $<MASK>$ picture taken."

\subsection{The Fill-Mask Model}

Mask-language modeling refers to bidirectional models like Bert \citep{devlin2018bert}, which handle tokens of texts from both left-to-right and right-to-left in order to conceptualize a sentence or phrase structure. Masking words refers mainly to the language model's inner ability to hide and predict missing text parts. Models like BERT were trained on large amounts of unsupervised text through \textit{missing and predicting}, which is nothing more than randomly hiding words in sentences to teach the model to predict them. Given that this is a well-grained ability of models like BERT, researchers do not need large amounts of labeled data to train it on downstream tasks \citep{icaart23}. In this sense, a plethora of out-of-the-box models can be used or fine-tuned in similar tasks. Given that we would always want our model to return pronouns for masked text, we use a specified pre-trained model from the Hugging Face platform. This refers to a RoBerta-base model trained on 10B of tokens one the task of predicting missing pronouns from sentences (see \textit{Fill-Mask Model} in Figure \ref{methodologyfig}). As this is one of the first works ever, we assume this pre-trained model to suffice for our experiments. Fine-tuning a pre-trained bidirectional model from scratch is out of the reach of this paper.

From this point on, and for every masked pronoun position, the model parsers each sentence to output a list of two pronouns (called top\_k) re-ordered by their significance, where the higher the percentage, the higher the possibility for the pronoun to be the most appropriate one ---although top\_k could be equal any number, number two was chosen for simplicity reasons. It should be noted that the model does not have access to the masked pronouns, meaning that it is likely, at some point, to return as an answer a hidden pronoun itself. Moreover, there is no a priori reason for assuming that every returned pronoun will have a gender matching with its referent or, under any given circumstances, to be the most appropriate one. In this regard, a more fine-grained procedure occurs as a plugged-in mechanism.

\subsection{Merger}
Before we dive into our most crucial part, a Merger component occurs. This simplified tool amasses the results of the Identifier component's sentences and the fill-mask models' results (see \textit{Merger} in Figure \ref{methodologyfig}). To illustrate, take, for example, a given sentence (s) with several pronouns (p). Knowing that our fill-mask model always returns two pronouns for each masked position (top\_k=2), this would always result in p*2 sentences (called siamese sentences). For instance, for the sentence ``The cat looked at the big dog, and $<mask>$ was terrified", knowing that the model would return two possible pronouns for the masked position (it, he), this would result in the following siamese sentences: i) The cat looked at the big dog, and $<it>$ was terrified .", ii.) "The cat looked at the big dog, and $<he>$ was terrified".

\subsection{The Winventor Component}
According to \citet{icaart20,isaak2021blending}, Winventor is a machine-driven approach for developing Winograd schemas, that is, sentence pairs (halves), where the objective is to resolve a definite pronoun to one of two co-referents in each half. Given that systems developed to tackle the WSC have typically been evaluated on a small set of hand-crafted schemas and that there is a plethora of developed systems that need to be evaluated on new schemas, Winventor comes to cover the need for an extensive and presumably continuously replenished, collection of available Winograd Schemas.

In this sense, Winventor uses a combination of tools to enhance the schema development process from sentences found on the English Wikipedia based on various controls that ensure quality. If it cannot develop a schema, it only develops the first half, consisting of the sentence, the question, and two pronoun targets. To illustrate, for the sentence ``The cat caught the mouse because it was clever.", Winventor could produce the next schema: [1st Half: Sentence: The cat caught the mouse because it was clever. Question: Who is clever? Answers: The cat, The mouse], [2nd Half: Sentence: The cat caught the mouse because it was careless. Question: Who is careless? Answers: The cat, The mouse]. 

In a lucid and approachable way, Winventor utilizes tools for spelling and grammar correction and finds semantic relations between entities and pronouns to select the best pair in each sentence, which is advantageous for our current work. It first utilizes spaCy to output semantic relations from a given sentence to locate the candidate pronoun targets and the events they participate in. Second, it searches for pronoun-noun and pronoun-proper-noun relations to develop extended relations, which refer to triples or scenes that indirectly relate various entities based on pronouns found in specified sentences \citep{Isaak2016}. It is worth mentioning that the pronoun targets have to belong to the same gender and be either singular or plural in order to meet the challenge rules.

Proper nouns are selected through spaCy’s entity recognition system, while nouns are filtered by pre-downloaded gender lists for gender agreement. Recognizing that in the English language, there is not always a gender agreement between nouns and pronouns, Winventor works in two modes: austere and broad-flag. In the former mode, there is always a gender agreement between entities and pronouns, while in the latter, there is no prerequisite for this. To maximize its efficiency, Winventor scores its half or schema based on several factors, where the higher the score, the more valuable the result is. Among others, factors like numberAgreement, genderAgreement, and pronounGenderAgreement, take the value one if they are true. Otherwise, they are set to zero. Pronouns like he/him/his refer to masculine entities, she/her(s) to feminine, it/its to neuter, and finally, they/them/their(s) are neutral ones. Moreover, there is an extra factor referring to the \citet{mitkov1998robust} score, which relates to five indicators that play a definitive role in hunting down the best pronoun targets from a set of possible candidates. In short, 1.) Definiteness refers to definite nouns that get a higher score than others. 2.) Indicating-Verbs list specialized verbs that precede important noun phrases. 3.) Lexical-Reiteration points out that repeated synonymous noun phrases are preferred over others. 4.) Non-Prepositional indicates that non-prepositional noun phrases should always get a higher preference than prepositional ones. 5.) Collocation points out that pronoun targets and pronouns of identical collocation patterns must always get a higher preference than other ones. Based on all the above, a specialized scoring mechanism is applied. Moreover, since more than one pair of nouns might exist in sentences, for any given sentence with more than two nouns, Winventor builds multiple schemas. More on how Winventor works can be found in the papers it was introduced with \citep{icaart20,isaak2021blending}.

In this work, we utilize Winventor to propel the selection of the most appropriate pronouns in sentences (see \textit{Winventor} in Figure \ref{methodologyfig}). To that end, via Winventor, we start parsing every given sentence from the Merger component (see \textit{Merger} in Figure \ref{methodologyfig}). Therefore, we leverage Winventor's ability to pair nouns with definite pronouns in order to evaluate sentences that match certain criteria related to factors mentioned above ---numberAgreement, genderAgreement, pronounGenderAgreement, and Mitcov score. Considering that our sentences might include multiple pronouns with more than two pronoun targets, we probe Winventor's ability to design multiple schemas to the next so-called suggestion level, that is, showing pairs of nouns and pronouns with a gender-like agreement along with their Mitcov scores.

\subsection{Matcher}
Matcher is plug and play component designed to estimate the appropriateness of each Siamese sentence based on values from both the Fill-Mask model and Winventor's component. Towards that end, each sentence is categorized based on various factors, such as its pronoun targets genderAgreement, pronounGenderAgreement, numberAgreement, Mitkov's score, and finally, based on the model's pronoun scores. Given that many tests can be answered in different ways, Matcher, through various controllable variables, can return sentences based on the factors above (see \textit{Matcher} in Figure \ref{methodologyfig}).

\begin{figure*}[h!]
	\centerline{\includegraphics[width=\textwidth]{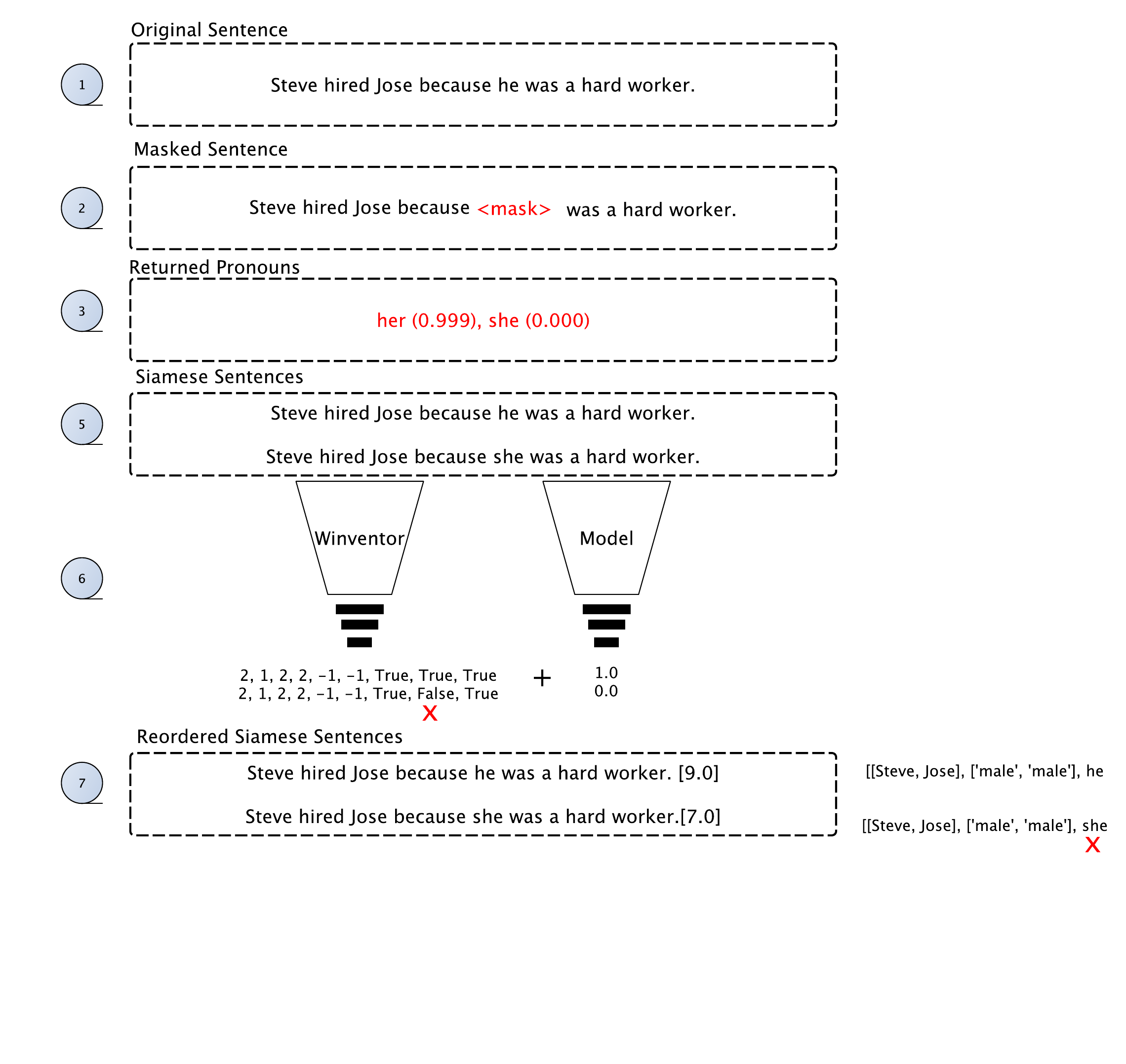}}
	\caption{An example of Matcher's Selection Process: Matcher reconstructs sentences based on returned pronouns to select the best one according to values returned by Winventor. The original sentence was taken from \citet{kn:Rahman:2012:RCC:2390948.2391032} study.}
	\label{maskSelection}
\end{figure*}

\section{Experimental Evaluation}
This section demonstrates PronounFlow's ability to select the most appropriate pronouns in sentences. In this regard, here, we present results obtained by applying the methodology described in the previous sections. Experiments run on a web-based Jupyter IDE notebook based on a RoBERTa language model\footnote{https://huggingface.co/thefrigidliquidation/roberta-base-pronouns} that runs on a dedicated GPU from the Gradient Paperspace platform\footnote{https://www.paperspace.com}.  

For evaluation purposes, we use sentences from Wikipedia, the Definite-Pronoun-Resolution dataset (DPR) \citep{kn:Rahman:2012:RCC:2390948.2391032}, and sentences found on specific WWW websites. The Wikipedia sentences refer to sentences downloaded from the English Wikipedia, used in developing Wikisense, a commonsense and reasoning system that tackles Winograd schemas \citep{Isaak2016,isaak2021blending}. On the other hand, the DPR dataset, consisting of 943 schemas, was manually developed by thirty undergraduate university students for an undertaken study by \citet{kn:Rahman:2012:RCC:2390948.2391032}. Each schema consists of pairs of halves with a sentence, a definite pronoun, two possible pronoun targets, and the correct pronoun target. Finally, sentences from the WWW refer to a selected set of sentences built around neopronouns.

Below, we present the results of two studies we undertook to evaluate PronounFlow's performance. In the first experiment, we tested our system on replicating sentences from the specialized DPR dataset. For evaluation purposes, every PronounFlow-parsed sentence is compared to its original version based on a \textit{hit and miss} approach to output our systems successfulness, that is, a percentage value showing the number of times PronounFlow successfully proposed all of the original pronouns. Given that sentences from the DPR dataset were thoroughly constructed for the WSC, we repeated the same experiment and compared the results to randomly selected sentences from the English Wikipedia. 

For concreteness, we also tested PronounFlow's a priori appropriateness as a helper tool for better coreference resolution. In this neologism pronoun experiment, we randomly selected sentences with epicene pronouns from the WWW and tested Stanford CoreNLP's success on coreference resolution prior to and post-PronounFlow's parsing procedure \citep{manning2014stanford,Isaak2016}.

\subsection{Results on Replicating Pronouns}
Here we tested PronounFlow's ability to replicate pronouns in well-structured sentences initially designed for the Winograd Schema Challenge. Consequently, 943 sentences from the DPR dataset, each from the first part of each schema, were parsed and analyzed with PronounFlow according to our proposed methodology. Recall that a group of Siamese sentences is returned for each examined sentence, accordingly reordered by their accumulated score based on gender, number, pronoun-gender, Mitcov, and language model values. For this experiment, the first three values equal one if they are true, otherwise, zero (see an example in Figure \ref{maskSelection}).

According to our results, PronounFlow parsed 914 sentences and rejected 29 whose pronoun was not in the language model's fine-tuned set (e.g., the pronoun him\footnote{\url{https://huggingface.co/thefrigidliquidation/roberta-base-pronouns}}). It seems that PronounFlow successfully identified the correct pronoun in 90\% of the cases, showing that our theory copes equally well with our experimental results. Moreover, an undertaken qualitative analysis revealed an average sentence length of 14.9 words, with an average number of 1.10 pronouns.  

Further experiments regarding the type of PronounFlow's aggregation score showed that in 52\% of cases, it was primarily based on Winventor values, while the rest, 48\%, was based on the model predictions. The results indirectly suggest that finding matching relations between pronouns and entities is a challenging and troublesome task as it depends on the kind of nouns and proper nouns used. Nevertheless, we must keep in mind that the accuracy results do not change whatever the mechanism used, Winventor or language model based. Of course, given that we are always interested in transparent and not opaque results, the process always starts with the Winventor component and only continues with the language model results in case the former fails. Moreover, PronounFlow's innate mechanism is considered as one based on the fact that a sentence with not matching relations always gets an arbitrarily added Winventor value of -60 to push it to the last positions of each Siamese group.

Considering that all DPR sentences were explicitly designed for the WSC, meaning that each sentence's pronouns and entities were thoroughly chosen, we repeated the same experiment with randomly selected chunks of Wikipedia sentences \citep{Isaak2016} ---for the sake of simplicity, our experiments are limited to chunks of 200 sentences. Through this variation, we wanted to see if PronounFlow could correctly identify the most appropriate pronouns in raw sentences designed for reasons other than pronoun resolution challenges. 

The results yielded some interesting findings favoring our engine's symbolic part. Our results indicate an association between Winventor's usage and PronounFlow's accuracy, showing the symbolic part as the optimal way to calibrate pronouns in sentences. Although each part comes with its pros and cons, undertaken experiments with randomly selected sentences of different lengths showed varied results from 77\% to 88\% in favor of Winventor's components. Please note that, based on our experiments and the utilized sentences, Winventor's part was measured from as low as 14\% to as high as 30\%, as it directly depends on every sentence's structure (see Table \ref{replicatingtwo}). PronounFlow seems to achieve its goals better when a sentence's entities are recognized, categorized, and matched to pronouns. Furthermore, these indications are reinforced by our previous experiments, where our system was more successful in identifying and matching pronouns with entities in well-structured sentences designed for the WSC. 

Altogether, our findings provide evidence of PronounFlow's ability to select and replicate pronouns in various sentences, even though, under certain circumstances, their structure can throw a wrench in the works of our system. Subsequently, considering that with a limited number of words, we could have an unlimited number of sentences, our results should be taken with a grain of salt as, at present, we are not in a position to determine if they can be applied to every type of sentence \citep{isaak2023neuralsymbolic}. Nevertheless, considering our accuracy results, all point to a well-coping mechanism of PronounFlow's symbolic and subsymbolic parts. 

\begin{table}[h]
	{\caption{A Screenshot of Applying PronounFlow On Randomly Selected Wikipedia Sentences of Various Lengths: The average sentence length in the first column is displayed, while in the second, what led PronounFlow to its conclusions. Finally, in the third column, we can see PronounFlow accuracy results in replicating the original pronouns.}\label{replicatingtwo}}
	\begin{adjustbox}{width={\columnwidth}}
\begin{tabular}{|c|c|c|}
	\hline 
	\multirow{2}{*}{Average Sentence Length} & Model - Winventor & \multirow{2}{*}{Score}\tabularnewline
	\cline{2-2} 
	& (Participation) & \tabularnewline
	\hline 
	11 & 0.86 - 0.14 & 0.77\tabularnewline
	\hline 
	14 & 0.78 - 0.22 & 0.81\tabularnewline
	\hline 
	18 & 0.74 - 0.26 & 0.86\tabularnewline
	\hline 
	22 & 0.71 - 0.29 & 0.88\tabularnewline
	\hline 
\end{tabular}
	\end{adjustbox} 
\end{table}

\begin{figure*}[h!]
	\centerline{\includegraphics[width=\textwidth]{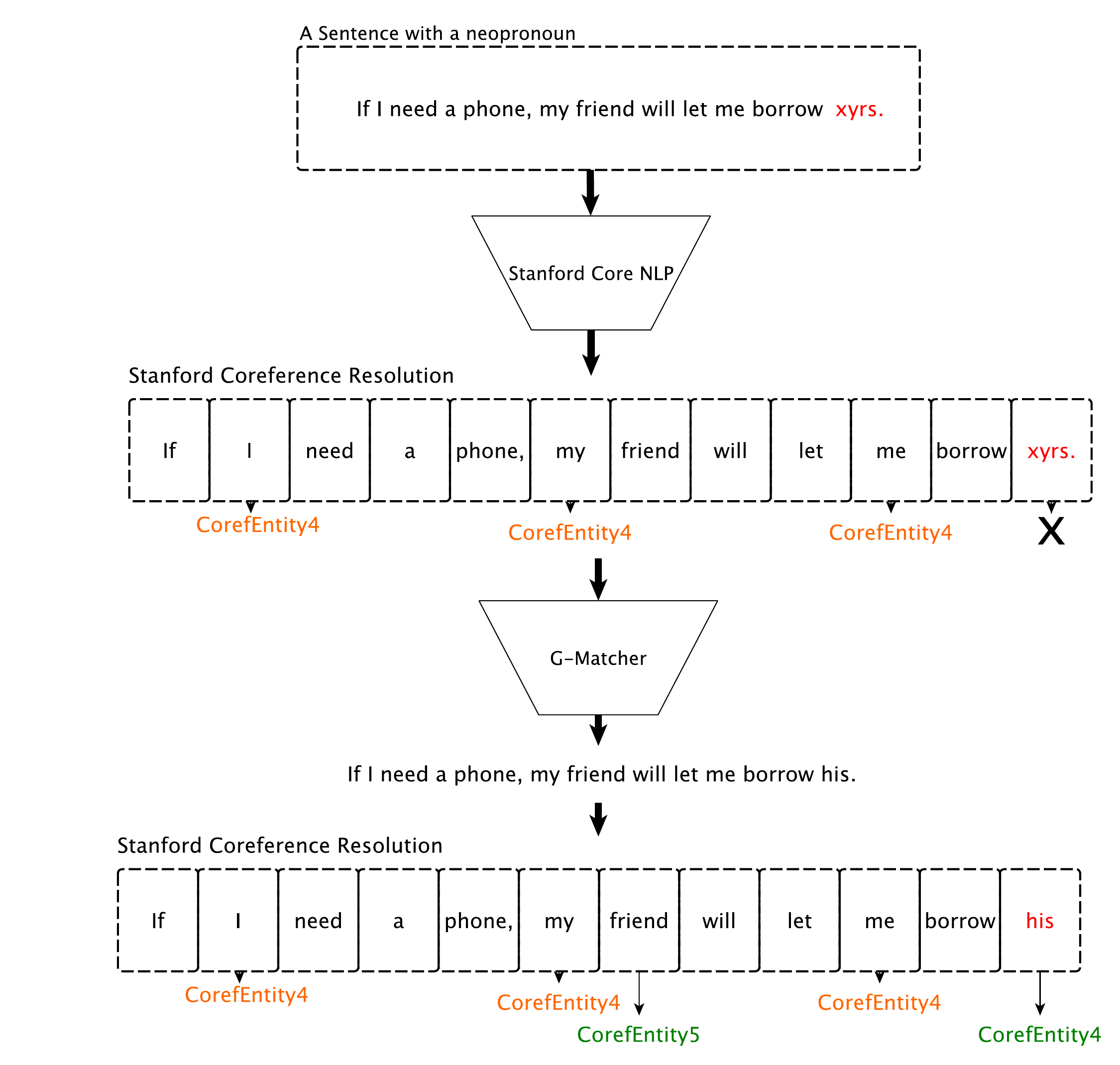}}
	\caption{An Example of Stanford Core NLP Coreference Resolution on a Sentence with \textit{Neopronouns}, prior and post PronounFlow's Application Process. Prior PronounFlow, Stanford was unable to find the appropriate antecedents of neopronouns.}
	\label{experiment2}
\end{figure*}

\begin{table}[h]
	{\caption{A Screenshot of Applying PronounFlow On Sentences with Neopronouns: On the first four rows, we can see sentences prior, while on the last four post-PronounFlow's parsing. Note that, for each examined sentence, PronounFlow can be programmed to return more than one result.}\label{coreferenceNeoExamples}}
	\begin{adjustbox}{}
	\begin{tabular}{|c|}
		\hline 
		With Neopronouns (Prior PronounFlow)\tabularnewline
		\hline 
		\hline 
		Sie is fun. Sie loves herself.\tabularnewline
		\hline 
		Xyr eyes grew wide.\tabularnewline
		\hline 
		If I need a phone my friend will let me borrow zirs .\tabularnewline
		\hline 
		I spoke with ver.\tabularnewline
		\hline 
		\multicolumn{1}{c}{}\tabularnewline
		\hline 
		Without Neopronouns (Post PronounFlow)\tabularnewline
		\hline 
		She is fun. She loves herself\tabularnewline
		\hline 
		Her eyes grew wide\tabularnewline
		\hline 
		If I need a phone my friend will let me borrow his .\tabularnewline
		\hline 
		I spoke with her.\tabularnewline
		\hline 
	\end{tabular}
	
	\end{adjustbox} 
\end{table}

\subsection{Results on Coreference Resolution}
This experiment provides empirical evidence of PronounFlow's ability to help state-of-the-art coreference resolution systems disambiguate pronouns in certain sentences (see Table \ref{coreferenceNeoExamples}). 

Specifically, here we examine hard cases of pronoun resolution within epicene sentences, that is, sentences built around neopronouns, and compare the results to siamese ones, that is, calibrated ones with post-PronounFlow's parsing procedure. Showing an improvement in the coreference resolution of Siamese sentences would suffice to show the importance of PronounFlow in the coreference resolution scene. 

In this regard, in the first step, we collected sentences with neopronouns from around the WWW. An initial attempt to collect large amounts of sentences from sources like the English Wikipedia was unsuccessful, indirectly showing the limitations in finding sentences with neopronouns. Ultimately, we collected 76 sentences built on top of fifty-nine neopronouns such as xe, ey, ze, xy, xyr, zirself, etc.

Next, prior PronounFlow, each sentence was parsed through Stanford CoreNLP\footnote{\url{https://corenlp.run}}. According to the results, prior PronounFlow, Stanford CoreNLP, failed to resolve even a single neopronoun. Moreover, from the 76 sentences, it only identified some entities in seven of them, albeit without any pronoun resolution. However, post-PronounFlow's parsing, Stanford CoreNLP correctly resolved pronouns and identified entities in all sentences, showing the importance of incorporating such tools in the coreference resolution process (see an Example in Figure \ref{experiment2}). 

Of course, on the flip side of the view, when dealing with new pronouns in the scene, some might argue against PronounFlow's suggestions. In this sense, we argue that our results should be taken with a grain of salt as they can be subjective. However, given that PronounFlow can be programmed to return more than one Siamese sentence for each example and considering the problems current coreference resolution systems seem to face, the results ultimately show that the blending of tools like PronounFlow and  Stanford CoreNLP can considerably help the latter in the pronoun disambiguation process. On a second front, developing tools like PronounFlow might become the precedent prior to any coreference resolution systems to remove or mitigate barriers to comprehension skills or even to help coreference resolution systems and readers construct inferences based on easily available information \citep{corbett1983pronoun,gernsbacher1990investigating}. As far as we can determine, tools like PronounFlow shed light on the potential of a new pronoun disambiguation era, where existing coreference resolution systems utilize them to lessen their drudge work.

\section{Conclusion}
We have introduced PronounFlow, a robust coreference resolution spin-off and a pronoun calibrator for English Sentences. Given a sentence, PronounFlow analyzes its structure to locate and replace pronouns that do not have a gender agreement with their entities or are not commonly used by state-of-the-art language models. To do that, it utilizes tools such as dependency parsers, language models, and components built for the WSC, able to match entities with their pronouns. Through extensive experiments, we showed the feasibility of our approach, suggesting that PronounFlow can successfully parse sentences to filter out unmatched pronouns. The ability to select, mask, and propose pronouns for replacement is done through language models, indirectly showing their ability to memorize and successfully predict missing words from any given text.

Our findings suggest that tools like PronounFlow can calibrate pronoun usage to design sentences according to the collective human knowledge found around us. On a second front, given that better pronoun resolution leads to a better understanding of the causal events in text passages, our work indirectly pushes forward a better pronoun resolution of existing coreference tools.

Future work can continue to explore pathways towards a better pronoun calibration process. This can be achieved by enhancing the Winventor component's gendered lists or completely replacing them with out-of-the-box fine-tuned language models. Moreover, one could try to shed light on PronounFlow's capabilities in parsing sentences of different types, such as simple, compound, complex, and compound-complex sentences.

\bibliographystyle{apalike}
{\small
	\bibliography{isaak}}

\end{document}